\documentclass[letterpaper, 10pt, journal, twoside]{IEEEtran}

\IEEEoverridecommandlockouts                              

\usepackage{graphicx} 
\usepackage{times} 
\usepackage{amsmath} 
\usepackage{amssymb}  
\usepackage{url}
\usepackage{cite}
\usepackage{enumerate}
\usepackage{bm}
\usepackage{cases}
\usepackage{mathtools}

\usepackage{multirow}
\usepackage{colortbl}
\usepackage{booktabs} 
\usepackage{makecell}
\usepackage{algpseudocode}

\usepackage[caption=false, font=footnotesize]{subfig}
\usepackage{algorithmicx}
\usepackage{algorithm}
\usepackage{algpseudocode}
\usepackage{booktabs, ragged2e}
\usepackage{tabularx}
\usepackage{xcolor}
\usepackage[para]{threeparttable}
\usepackage{colortbl}

\makeatletter
\let\NAT@parse\undefined
\makeatother
\usepackage[hidelinks]{hyperref} 

\DeclarePairedDelimiter\abs{\lvert}{\rvert}%

\newcolumntype{C}{>{\Centering\arraybackslash}X}
\newcolumntype{L}{>{\raggedright\arraybackslash}X}
\newcolumntype{R}{>{\raggedleft\arraybackslash}X}

\algnewcommand\algorithmicnot{\textbf{not}}
\algdef{SE}[IF]{IfNot}{EndIf}[1]{\algorithmicif\ \algorithmicnot\ #1\ \algorithmicthen}{\algorithmicend\ \algorithmicif}%

\begin{document}
\title{Efficient Speed Planning for Autonomous Driving in Dynamic Environment with Interaction Point Model

\author{
\begin{tabular}{@{}c@{}}
Yingbing Chen, 
Ren Xin,
Jie Cheng, \textit{Student Member, IEEE}, \\ 
Qingwen Zhang, \textit{Student Member, IEEE},
Xiaodong Mei, \textit{Student Member, IEEE}, \\
Ming Liu, \textit{Senior Member, IEEE},
and Lujia Wang, \textit{Member, IEEE}
\end{tabular}}

\thanks{© 20XX IEEE.  Personal use of this material is permitted.  Permission from IEEE must be obtained for all other uses, in any current or future media, including reprinting/republishing this material for advertising or promotional purposes, creating new collective works, for resale or redistribution to servers or lists, or reuse of any copyrighted component of this work in other works}


\thanks{Authors are with Robotics Institute, the HKUST, Hong Kong SAR, China. (e-mail: ychengz@connect.ust.hk, rxin@connect.ust.hk, jchengai@connect.ust.hk, qzhangcb@connect.ust.hk, xmeiab@connect.ust.hk, eelium@ust.hk, eewanglj@ust.hk. (\textit{Corresponding author: Ming Liu.})}%

\thanks{Digital Object Identifier (DOI): 10.1109/LRA.2022.3207555.}
}

\markboth{IEEE Robotics and Automation Letters. Preprint Version. September, 2022}
{Chen \MakeLowercase{\textit{et al.}}: Efficient Speed Planning with Interaction Point Model} 

\maketitle

\begin{abstract}

Safely interacting with other traffic participants is one of the core requirements for autonomous driving, especially in intersections and occlusions.
Most existing approaches are designed for particular scenarios and require significant human labor in parameter tuning to be applied to different situations.
To solve this problem, we first propose a learning-based Interaction Point Model (IPM), which describes the interaction between agents with the \textit{protection time} and \textit{interaction priority} in a unified manner. 
We further integrate the proposed IPM into a novel planning framework, demonstrating its effectiveness and robustness through comprehensive simulations in highly dynamic environments.

\end{abstract}

\renewcommand{\arraystretch}{1.0}
\section{Introduction\label{sect:intro}}

\begin{figure}[!t]
    \centering
    \includegraphics[width=3.0in]{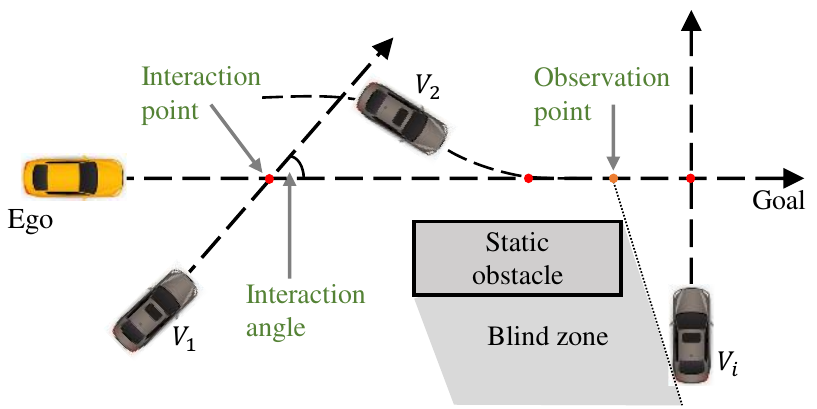}
    \caption{The AV (ego) follows the path to the goal with considerations of interactions with other traffic agents (e.g., $V_i$) at different scenarios (different interaction angles and the blind zone).}
    \label{fig:overview}
\vspace{-1.5em}
\end{figure}

\IEEEPARstart{P}{lanning} safe trajectories for autonomous vehicles (AVs) in complex environments is a challenging task. As demonstrated in Fig. \ref{fig:overview}, it requires the planning system to have the ability to safely react in real-time under various scenarios, such as intersection, merging, and traversing areas with a limited field of view (FoV). 

A typical strategy in solving these problems is to transform traffic rules and prediction results into constraints in an optimization problem \cite{zhan2017spatially, gao2018online}. Although this method can generalize to different scenarios, it does not model the prediction and interaction uncertainties well.
Another strategy applies the POMDP-based methods\cite{hubmann2018automated, zhang2020pomdp_efficient}, which searches for the best reaction trajectory by simulating the traffic interactions in the belief space. However, these methods introduce a high computational burden,
and are hard to reflect the interaction patterns using manually designed reward functions.

\begin{figure}[!t]
    \centering
    \subfloat[Demonstration of $\Delta \theta$ and $\Delta t$]{
        \begin{minipage}[t]{0.5\linewidth}
            \includegraphics[width=\linewidth]{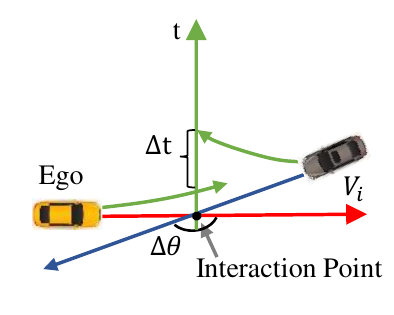}
        \end{minipage}
        \label{fig:illus_of_protection_dt}}
    \subfloat[Area with limited FoV]{
        \begin{minipage}[t]{0.5\linewidth}
            \includegraphics[width=\linewidth]{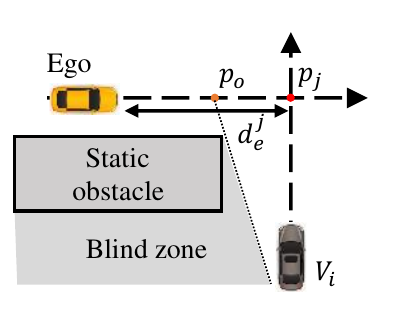}
        \end{minipage}
        \label{fig:occlusion_speed_limit}}

    \caption{Illustrations of interactions in traffic. (a) Demonstration of the interaction angle $\Delta \theta$ and the time gap $\Delta t$, where the AV goes along the red path and $V_i$ moves along the blue one. (b) The AV can not determine its maneuver until it reaches the observation point $p_o$.}
    \label{fig:illus_details}
\vspace{-1.5em}
\end{figure}

To better capture the interaction among traffic participants, it is essential to focus on modeling driving patterns around an ``interaction point" without traffic regulations. The interaction points are observable situations where at least two road users intend to occupy the same space in the near future\cite{2020interaction_defining}.
Some works implicitly model the planning problem on these points. For instance, car-following models\cite{koopman2019RSS, makridis2019response} and lane-merging approaches\cite{jeong2019target} use different values of time gap ($\Delta t$ in Fig. \ref{fig:illus_of_protection_dt}) to generate the proper reactive behavior for AVs. Prediction work\cite{sun2022domain} analyzes the right-of-way of collision points in the intersection. However, these methods are tailored for specific scenarios. It brings a heavy load on parameter tuning when applying multiple methods in different situations. 

When reacting to other road users at different speed and angles in various scenarios, human drivers generally show different degrees of caution that can be indicated by the time gap. Therefore, we focus on the following two questions. \textbf{Question 1}: When two agents intend to pass the same interaction point in various scenarios, what is the minimum time gap between them? 
\textbf{Question 2}: What determines the priority of agents to pass the interaction point in \textit{Question 1}?

To answer these questions, we propose a novel Interaction Point Model (IPM), where \textit{Question 1} is solved by statistical analysis in real driving records, and \textit{Question 2} is explained using a multilayer perceptron (MLP) network.
Then, based on IPM, we present a unified planning framework\footnote[2]{IPM model files and demonstration videos are available on the project website: \href{https://github.com/ChenYingbing/IPM-Planner}{\color{blue}{https://github.com/ChenYingbing/IPM-Planner}}.} to achieve safe and interactive motion planning, which guides the ego vehicle pass an interaction point only when it has a higher priority than other traffic participants.

\subsection{Contributions\label{subsect:contribution}}


Overall, we propose a unified interactive planning framework for AVs. The major contributions are:
\begin{itemize}

    \item A novel and general interaction solution called IPM, which uniformly formulates the interaction in terms of the protection time and priority. 
    

    \item A unified planning framework for AV in complex dynamic environments, including an efficient s-t graph searching to calculate the trajectory distribution under traffic regulations, and a priority determination module based on the IPM to select a safe trajectory.


    \item Evaluations in unsignalized urban driving simulations showing our method's computational efficiency, robustness, and effectiveness.
\end{itemize}

\section{Related Work}


\subsection{Interaction Representation}
Existing methods used different ways to represent traffic interactions in planning. One strategy was to represent prediction results and simulation outcomes\cite{zhang2020pomdp_efficient, hanna2021pomdp} in the spatio-temporal (s-t) graph. Then, the sampling methods\cite{lim2019hybrid, zhang2020trajectory} and optimization methods\cite{ding2019safe} were used to extract the best-cost or lowest-risk trajectory.
Another strategy applied reactive or interactive models\cite{hubmann2018automated} to generate action for AVs directly. Methods based on elaborated rules\cite{koopman2019RSS, zhang2016combined}, game theory\cite{rahmati2021helping}, and control theory\cite{li2021prediction} were widely used in autonomous driving.
Compared to these methods, our work applies both s-t graph representation and interactive model (IPM). This combination makes it avoid deterministic representation\cite{lim2019hybrid, ding2019safe, jie_icra2022} and limited expression of interaction\cite{koopman2019RSS, zhang2016combined, rahmati2021helping, li2021prediction}. In addition, our method is highly efficient as it does not require low-efficient simulation\cite{zhang2020pomdp_efficient, hanna2021pomdp} and accurate collision checking\cite{lim2019hybrid}. It also outperforms learning-based methods\cite{cj2022mpnp} on interpretability, as the IPM is simpler and more intuitive.

\subsection{Interactive Motion Planning for AVs}
In interactive planning, the motion of other traffic agents and the ego vehicle is modeled as interdependent. This interdependency makes it impracticable to separate the problems of prediction and planning\cite{hubmann2018automated}. There were several interactive planning approaches. 
For the discussed POMDPs\cite{hubmann2018automated, zhang2020pomdp_efficient, hanna2021pomdp}, interactions were reflected by time-consuming simulations but not modeled directly. In recent years, learning-based approaches\cite{song2020pip, jiao2022tae} have been promising to learn trajectory generation and prediction in a unified architecture. However, machine learning technologies lacked traceability when an error occurred in practice. Another was game-theoretic plannings\cite{wang2021game, chandra2022gameplan}, aiming to compute the Nash equilibrium solutions for all players iteratively. The major shortage of these methods was their high computational cost when the number of players increased. In addition, the game rules were elaborately designed for specific situation, making them hard to generalize to different scenarios.

Besides these, branch MPC\cite{chen2022interactive} and fail-safe planners\cite{cui2021lookout_multi_modal, pek2018computationally, pek2020tro_fail_safe} emphasized the worst-case performance of AVs. For branch MPC\cite{chen2022interactive}, it was an optimization method over risk measures to minimize the worst-case expectation, which did not model the traffic interactions well.
Instead of accurate risk avoidance, fail-safe plannings \cite{pek2018computationally, pek2020tro_fail_safe} were reactive planners that guaranteed AV's safe reactions to all possible prediction uncertainties. It achieved this by keeping a collision-free maneuver available at all times. Following this idea, our method shows a conservative maneuver only when the interaction priority is estimated as low, which prevents the AV from too conservative behaviors\cite{pek2020tro_fail_safe}.

\section{Overview\label{subsect:overview}}

\begin{figure}[tb]
\centering
\vspace{6pt} 
\includegraphics[width=1.0\linewidth]{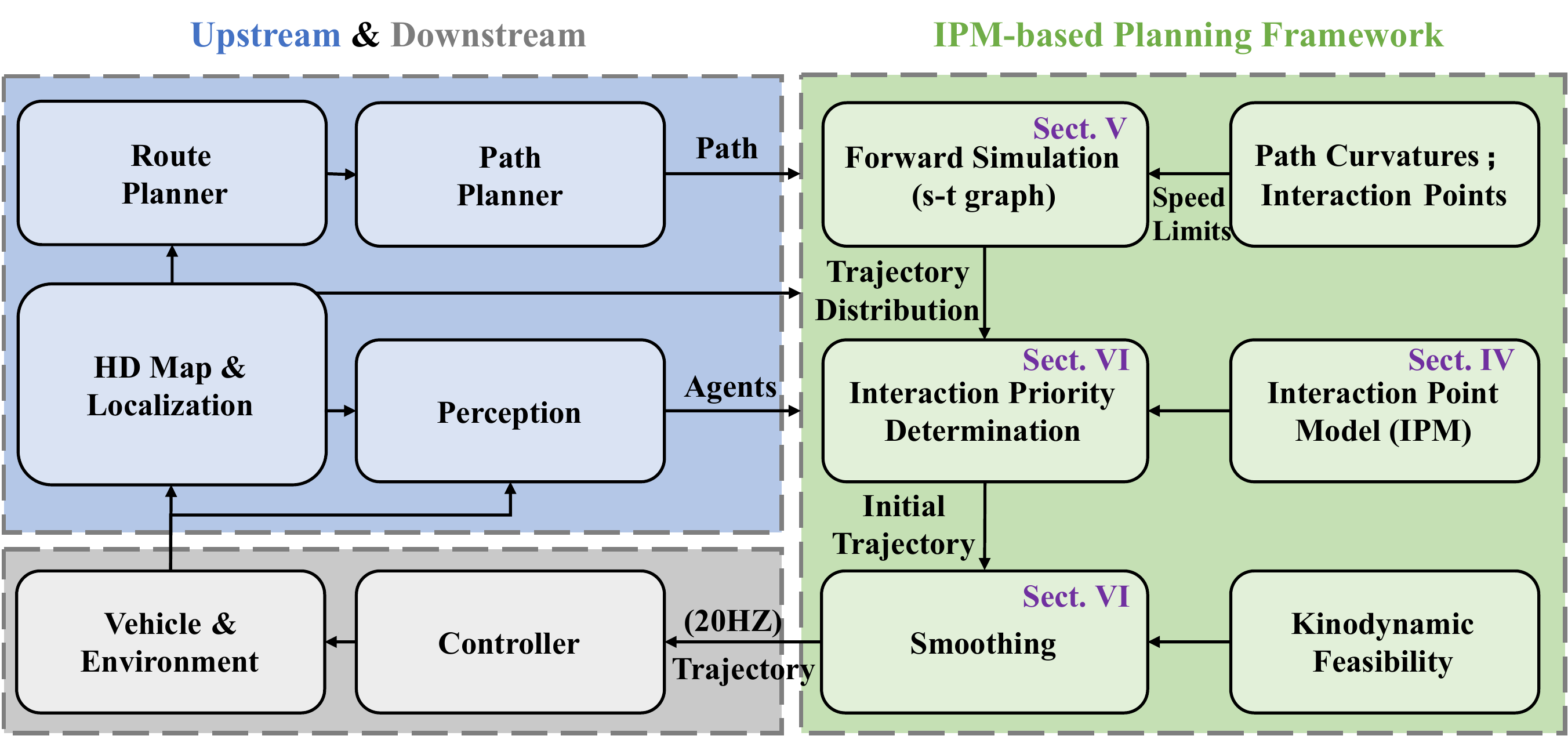}

\caption{Overview of the proposed planning framework with IPM and its relationship with other AV system components.}

\label{fig:overview-framework}
\vspace{-1.5em}
\end{figure}

An overview of the proposed planning framework is presented in Fig.\ref{fig:overview-framework}. In this system, the IPM has two functions: 1) determining whether the AV has a higher priority at an interaction point and 2) extracting the speed limit if the AV intends to pass an occluded area. The framework functions as follows: Given the inputs of the path and speed limits, the spatio-temporal (s-t) graph search first outputs the valid trajectory distribution. Then, the interaction priority determination decides a safe and efficient trajectory for the AV. Finally, the trajectory is smoothed to meet the kinodynamic feasibility and is delivered to the controller for execution.


\section{The Proposed Interaction Point Model\label{sect:def_ipm}}

\begin{figure}[tb]
    \centering

    \subfloat[Point-Overlap]{
        \begin{minipage}[t]{0.5\linewidth}
            \includegraphics{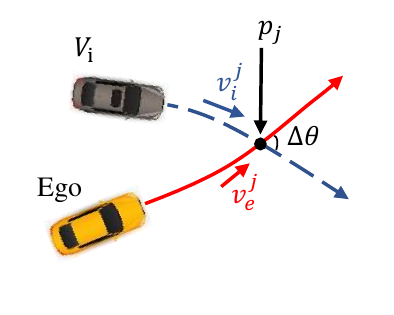}
        \end{minipage}
        \label{subfig:illus_point_overlap}}
    \subfloat[Line-Overlap]{
        \begin{minipage}[t]{0.5\linewidth}
            \includegraphics{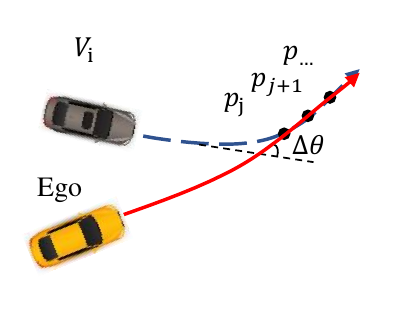}
        \end{minipage}
        \label{subfig:illus_line_overlap}}

    \caption{Illustrations of the interaction point $p$, angle $\Delta \theta$, and speed $v$. (b) shows that the interaction angle should consider the shapes of the agents.}
    \label{fig:illus_interaction_point}
\vspace{-1.5em}
\end{figure}

We extract pairs of interaction data from the INTERACTION dataset \cite{zhan2019interaction} to analyze the protection time and priority in the interaction points. Two examples are shown in Fig. \ref{fig:illus_interaction_point}, where car following is a special case of the line overlap \cite{zhan2017spatially}. It should be noted that the proposed IPM is also applicable in one-to-many interactions.

\subsection{Assumptions\label{subsect:ipm_assumption}}

The human-robot interaction model\cite{2006iteraction} indicates that participants will exchange their intentions beyond a certain distance called the social distance. Within the social distance, the determined form of interaction is performed. Similarly, \cite{2020interaction_defining} points out that, before an interaction happens, traffic participants show their intentions via implicit (interactive behavior) or explicit (e.g., turn signal) communications. Inspired by these works, we make the following assumptions. 

\textit{Assumption 1}: The participants will be conservative when facing potential interactions until the maneuver (e.g., who goes first) is determined via communication \cite{2020interaction_defining}. 

\textit{Assumption 2}: The maneuver for an interaction point can be determined only after all participants are observed. There is a location called an observation point at which the AV has no blind zone in FoV.

For \textit{Assumption 1}, we believe it is rational because even in a car-following scenario \cite{koopman2019RSS}, traffic agents act conservatively to keep a safe distance from the front vehicle. That is, the planned trajectory should guarantee the safety of the AV for the worst-case situations in the future \cite{pek2020tro_fail_safe}. As for \textit{Assumption 2}, an example of the observation point is shown in Fig. \ref{fig:occlusion_speed_limit}, where the AV cannot determine its maneuver to interact with the $V_i$ until it reaches the observation point $p_o$. How to obtain an observation point is not discussed in this work, it requires an analysis of the FoV and high-definition map \cite{lee2017blind_zone_collision}.

\subsection{Interaction Point \label{subsect:def_ip}}

As illustrated in Fig. \ref{subfig:illus_point_overlap}, $p_j$ denotes the $j$-th interaction point, and $v_e^j$ and $v_i^j$ are the speeds of the moment that agents pass the $p_j$. We further define the properties of the $p_j$ as
\begin{equation}
    \Delta v = v_e^j - v_i^j,
    \Delta \theta = |\theta_e^j  -\theta_i^j|,
    \Delta t = t_e^j - t_i^j, 
    \label{eq:interaction_dv_dangle}
\end{equation}
\noindent where $\Delta v$ and $\Delta \theta$ are the relative interaction speed and angle, and $\Delta t$ is the time gap in Fig. \ref{fig:illus_of_protection_dt}. In the equation, similar to the definition of $v_e^j$, $\theta_e^j$ is the yaw state of the AV at the location of $p_j$, and $t_e^j$ is the arrival time. The subscript letter indicates the agent, and the superscript corresponds to the index of the interaction point. Moreover, $\Delta t$ is nonzero because agents cannot pass the same position simultaneously. 

\subsection{Interaction Protection Time \label{subsect:interaction_protection_time}}

Following \textit{Question 1}, the interaction protection time is the minimum value of the time gap $\Delta t$. It has two cases: A negative $\Delta t$ corresponds to overtaking, and a positive one corresponds to yielding for others. As a result, we define
\begin{equation}
\left\{
\begin{aligned}
    \Delta t_{p}^- &= \max{\Delta t}, \forall \Delta t \in \mathcal{T} \cap \mathbb{R}^{-}, \\
    \Delta t_{p}^+ &= \min{\Delta t}, \forall \Delta t \in \mathcal{T} \cap \mathbb{R}^{+},
\end{aligned}
\right.
\end{equation}
\noindent where $\Delta t_{p}^-$ denotes the maximum time gap when overtaking, $\Delta t_{p}^+$ is the minimum time gap when giving way, and $\mathcal{T}$ is the set of possible values of the time gap in the data set. Then, $\Delta t_p^{(\cdot)}$ is modeled as functions of $\Delta \theta$ and $\Delta v$ respectively.
\begin{equation}
    \begin{aligned}
        \Delta t_p^-(\Delta v) &= \mathbf{C_1} 
        \begin{bmatrix}
            \Delta v^4 &
            \Delta v^3 & 
            \Delta v^2 & 
            \Delta v & 
            1 \\ 
        \end{bmatrix}^T, \\
        \Delta t_p^-(\Delta \theta) &= \mathbf{C_2} 
        \begin{bmatrix}
            \Delta \theta^2 & 
            \Delta \theta & 
            1 \\ 
        \end{bmatrix}^T, \\
        \Delta t_p^+(\Delta v) &= \mathbf{C_3} 
        \begin{bmatrix}
            \Delta v^4 &
            \Delta v^3 & 
            \Delta v^2 & 
            \Delta v & 
            1 \\ 
        \end{bmatrix}^T, \\
        \Delta t_p^+(\Delta \theta) &= \mathbf{C_4} 
        \begin{bmatrix}
            \Delta \theta^2 & 
            \Delta \theta & 
            1 \\ 
        \end{bmatrix}^T,
    \end{aligned}
\label{eq:fitting_curves}
\end{equation}

\noindent where $\Delta t_p^{(\cdot)}(\Delta v)$ and $\Delta t_p^{(\cdot)}(\Delta \theta)$ are represented as fitting curves, and $\mathbf{C_{(\cdot)}}$ is the fitting coefficient. 
Given the values of $\Delta v$ and $\Delta \theta$ together, the boundaries become

\begin{equation}
\left\{
\begin{aligned}
    \Delta t_{p}^- &= \min{\left(\Delta t_p^-(\Delta v), \Delta t_p^-(\Delta \theta) \right)}, \\
    \Delta t_{p}^+ &= \max{\left(\Delta t_p^+(\Delta v), \Delta t_p^+(\Delta \theta) \right)}.
\end{aligned}
\right.
\label{eq:protection_time}
\end{equation}


\noindent More illustrations will be given in Sect. \ref{subsect:experi_ipm_protect_time}.

\subsection{Speed Constraint at Observation Point \label{subsect:speed_cons_obs_point}}

Recalling \textit{Assumption 1} and \textit{2}, before all occluded vehicles are observed, the agent needs to prepare for the worst-case in the future, that is, to give way to any emergent vehicles. As the example shown in Fig. \ref{fig:occlusion_speed_limit}, before the AV reaches the observation point $p_o$, the remaining time $T$ to reach $p_j$ should be greater than $\Delta t_{p}^+$. This means the speed upper bound
\begin{equation}
\begin{aligned}
    &\overline{v} = \frac{d_e^j}{T} \leq \frac{d_e^j}{\Delta t_p^+}, d_e^j \geq \text{Dist}(p_o, p_j), \\
\end{aligned}
\label{eq:speed_cons_obs_point} 
\end{equation}
\noindent where $d_e^j$ is the distance between the AV and the $p_j$, $\text{Dist}(\cdot)$ returns the Euler distance, and the upper bound meets its minimum $\overline{v} = \text{Dist}(p_o, p_j) / \Delta t_p^+$. In this equation, $\Delta t_p^+$ can be quickly estimated by setting $\Delta v$ to zero, which means there is no need to predict agents' speed in the occluded area.

\subsection{Interaction Priority \label{subsect:interaction_priority}}

\begin{figure*}[!t]
    \centering
    \subfloat[Expansion and local truncation]{\includegraphics[width=3.4in]{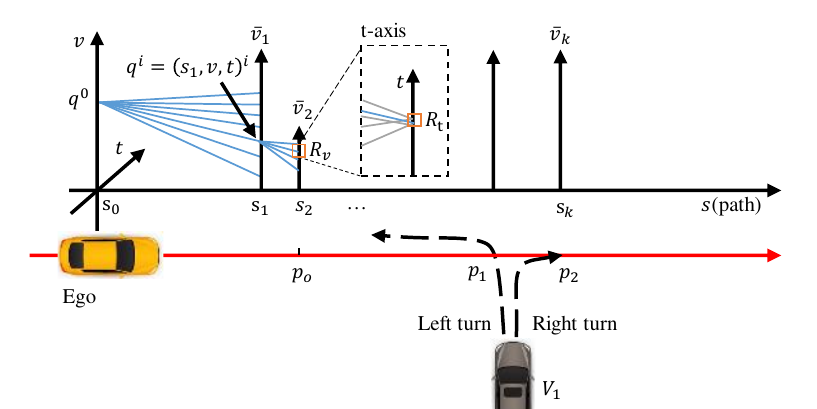}
    \label{subfig:expansion_and_local_trunc}}
    \hfil
    \subfloat[Speed profile generation]{\includegraphics[width=3.4in]{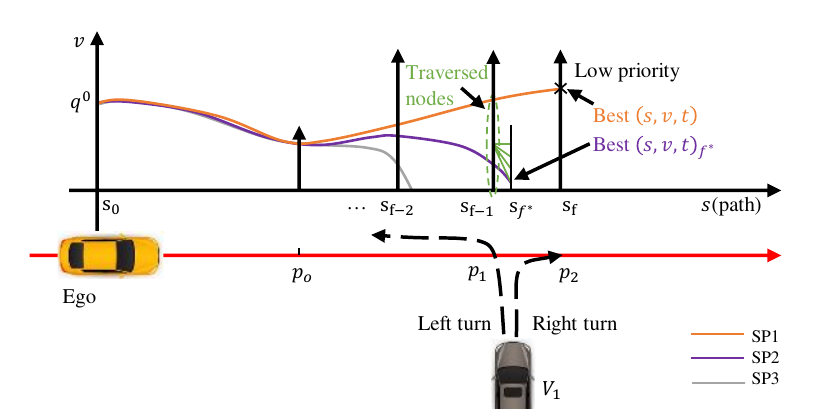}
    \label{subfig:speed_profile_gen}}

    \caption{Illustrations of the efficient s-t graph search and speed profile generation.
    (a) Results of the forward expansion for two rounds, where the grey lines in the t-axis are abandoned edges during local truncation.
    (b) Speed profiles (as colored lines) are generated in the interaction priority determination. 
    }
    \label{fig:st_graph_algorithm}
    \vspace{-1.0em}
\end{figure*}

The definition of the interaction priority is from \textit{Question 2}. It is treated as a conditional distribution $P_j(\Delta t<0 | \bm{f})$, where $\bm{f}$ is the feature vector. Before representing the input features, we introduce two metrics: the minimum arrival time $\min{(t_i^j)}$ and the maximum arrival time $\max{(t_i^j)}$. These two metrics are calculated by the constant acceleration model
\begin{equation}
\begin{aligned}
    \min{(t_i^j)} 
    &= \frac{\sqrt{{v_0}_i^2 + 2 \max(a_i) d_i^j} - {v_0}_i}{\max(a_i)},
\end{aligned}
\label{eq:train_eq1}
\end{equation}
\begin{equation}
\begin{aligned}
    \max{(t_i^j)} &= \frac{{v_0}_i - \sqrt{{v_0}_i^2 + 2 \cdot \min(a_i) d_i^j} }{|\min(a_i)|}, \\
    \text{s.t. } a_i &\geq -{v_0}_i^2 / (2 \cdot d_i^j),
\end{aligned}
\label{eq:train_eq2}
\end{equation}
\noindent where ${v_0}_i$ denotes the initial speed of agent $V_i$, 
$a_i$ is the possible average acceleration before reaching the interaction point, and $\max{(a_i)}$ and $\min{(a_i)}$ are the maximum and minimum values of acceleration. When the AV and $V_i$ intend to pass $p_j$, two features named the overtaking ability $M^-$ and giving way ability $M^+$ are defined as
\begin{equation}
\begin{aligned}
    M_j^- &= \min(t_e^j) - \min(t_i^j) + |\Delta t_p^-|, \\
    M_j^+ &= \max(t_e^j) - \max(t_i^j) - |\Delta t_p^+|, \\
\end{aligned}
\label{eq:train_eq_f}
\end{equation}
\noindent where the expression of $M^-$ indicates that the overtaking needs to suffer the protection time $\Delta t_p^-$, and $M^+$ is the time difference when the two agents slow down simultaneously. In this equation, when $M^-$ is less than zero, the AV can safely occupy the interaction point in advance because it can arrive much earlier than $V_i$. When $M^-$ is positive, if the AV wants to overtake, the communications mentioned in \textit{Assumption 1} and the traffic regulations must be considered.

Our goal is to learn a binary classifier $P_j(\Delta t < 0 | M^-, M^+)$. Since agents in the experiments are not as intelligent as in the dataset, the classifier is modified as
\begin{equation}
\left\{
\begin{aligned}
    P_j(\Delta t<0) &= 0.0, M^- \geq t_m, \\
    P_j(\Delta t<0) &= \textsc{MLP}(M^-, M^+), M^- < t_m,
\end{aligned}
\right.
\label{eq:ipm_priority}
\end{equation}

\noindent where \textsc{MLP} is the trained 3-layer MLP model, $t_m = \min(C_r^1\cdot t_e^j + C_r^2, C_r^3)$ is the time threshold, and $C_r^{(\cdot)}$ are constants ($C_r^1$ is positive). When the AV is far from the interaction point ($t_e^j$ is large), it is assumed that the interaction follows patterns learned from the dataset ($M^-$ is less than $t_m$). In contrast, when $t_e^j$ is small, it turns into a more conservative model in which $M^-$ plays a key role.

\section{Forward Simulation \label{sect:forward_graph_search}}

\begin{figure}[!t]
\vspace{-0.7em}
\begin{algorithm}[H]
\caption{S-t graph search}\label{alg:s-t_graph_search}
\begin{algorithmic}[1]
    \State \textbf{Notation}: open list $\mathcal{O}$, layer indexs $l$ and $f$, path points $\{s_1, s_2, ..., s_k\}$, node list $\mathcal{V}$, the maximum layer $k$, constants $t_h$ and $c_v$
    \State \textbf{Initialize}: $\mathcal{O} \gets \{q^0\}$, $\mathcal{V} \gets \emptyset$, $f \gets 0$
    \For{$n_{iter} \in \{0, \dots, \text{max\_iter}\}$}
        \State $q^i \equiv (s_l, v, t)^i \gets \textsc{PopBestNode}(\mathcal{O})$
        \State $f \gets \max(f, l)$, $\mathcal{V}_l \gets \mathcal{V}_l \cup q^i$
        \IfNot{\textsc{OutOfBounds}($q^i, l, k, t_h, c_v$)}
            \For{$a \in \{a_1,\dots,a_n\}$}
                \State $q^c \gets$ \textsc{ForwardExpansion}($q^i, a, s_l, s_{l+1}$)
                \State $J=$\textsc{CalculateCost}($q^i, q^c$)
                \If{\textsc{LocalTruncation}($q^c, J, R_v, R_t$)}
                \State $\mathcal{O} \gets \mathcal{O}\cup q^c$
                \EndIf
            \EndFor 
        \EndIf \algorithmiccomment{Check if not out of bounds}
    \EndFor
    \State $q^* \gets \textsc{BestNode}(\mathcal{V}_{f})$
    \State \textbf{return} $\mathcal{V}$, \textsc{PathTo($q^*$)}
\end{algorithmic}
\end{algorithm}
\vspace{-2em}
\end{figure}

The minimum arrival time in the IPM uses a constant acceleration model, which is not competent when the speed limit changes. Hence, a spatio-temporal (s-t) graph search method is applied to sample speed profiles along the path. This module does not generate a collision-free speed profile but obtains the future trajectory distribution under variational speed limits. 
As shown in Fig. \ref{fig:st_graph_algorithm}, we present the graph node as $q^i = (s_l, v, t)^i$, where $i$ is the node index, $s_{(\cdot)}$ is the accumulated distance, the subscript $l$ is called the layer index, and $v$ and $t$ are the speed and timestamp values. 



\subsection{Path Point Selection}

The first step is to select a sequence of path points $\{s_1, s_2, ..., s_k\}$ to build the search space, where $k$ is the maximum layer. This process considers the curvatures and locations of the interaction points, which obeys: (i) The selected point should reflect the speed limit change between bends and straights. (ii) The path point should involve the location of the interaction point and observation point. (iii) The number of layers should stay a suitable size. Too many layers will increase the computation amount, while a too-small number will reduce the search space.

\subsection{Speed Constraints \label{subsect:sped_cons_in_graph_search}}

During forward simulation, traffic elements, e.g., traffic lights, stop signs, and speed limits, can be transformed into spatio-temporal constraints in the search space \cite{ding2019safe}. This work involves two kinds of velocity constraints: one comes from the curvature, and the other from the observation point in (\ref{eq:speed_cons_obs_point}). We denote $a_{\text{lat}}^{\max}$ as the maximum lateral acceleration and $\kappa(s)$ as the curvature at $s$. The speed limit of the curvature \cite{jie_icra2022} is
\begin{equation}
    \overline{v}_{\kappa} \leq v_{\kappa}^{\max} =  
    \sqrt{
        a_{\text{lat}}^{\max} / \abs{\kappa(s)}
    }.
\end{equation}

\subsection{Forward Expansion \label{subsect:forward_expansion}}

We present the whole process in Algorithm \ref{alg:s-t_graph_search} and a demonstration in Fig. \ref{subfig:expansion_and_local_trunc}. During the expansion, the best node popped from the open list (Algo. \ref{alg:s-t_graph_search}, Line 4) is forwarded by
\begin{equation}
        v^c = \sqrt{
        (v^i)^2 + 2 a \cdot \Delta s}, t^c = t^i + \frac{v^c - v^i}{a},
\end{equation}
\noindent where $\Delta s = s_{l+1} - s_l$, and $q^c \equiv (s_{l+1}, v, t)^c$ is the child node. $v^c \in (\underline{v}, \overline{v})_{l+1}$, $(\underline{v}, \overline{v})_{l+1}$ are the speed bounds at layer $l+1$. $a$ is the acceleration from a set of discretized control inputs $\mathcal{U} = \{\underline{a} \leq a_1,\dots,a_n \leq \overline{a}\}$. During the expansion, any child node which exceeds the layer bound $k$ or time limit $t_h$, or has a speed that is lower than $c_v$ will not be expanded (Algo. \ref{alg:s-t_graph_search}, Line 6). The expanded node has cost
\begin{equation}
    J = J_p + w_a J_a + w_v J_v,
\end{equation}
\noindent where $J_p$ denotes the cost of the parent node, $J_a=u^2 \Delta t$ is the control cost, $J_v=|v - \overline{v}|$ is the deviation from the speed limit, and $w(\cdot)$ is the weights.

\subsection{Local Truncation \label{subsect:local_truncation}}

As illustrated in Fig. \ref{subfig:expansion_and_local_trunc}, for nodes in a discretized $(v, t)$ grid with resolutions $R_v$ and $R_t$, only the node with the lowest cost is allowed to undergo expansion (Algo. \ref{alg:s-t_graph_search}, Line 10). This operation significantly improves the search efficiency, and it has little impact on the search result since the abandoned nodes have similar values to the lowest-cost one.


\section{Interaction Priority Determination \label{sect:interation_prior_deter}}

Once the graph search is finished, trajectories to graph nodes $\mathcal{V}$ and the lowest-cost speed profile $q_{1:f} = \{q^1, q^2, \dots, q^f\}$ are obtained. For node $q^i$ in $q_{1:f}$, we use notation $i$ to represent both the node and layer index for convenience. The next step is to determine which trajectory to perform, and the IPM is applied for priority checking among the sampled trajectories.

\subsection{Speed Profile Selection}

The whole process is outlined in Algorithm \ref{alg:ipd_process}. $\mathcal{V}_{0:i}$ denotes the graph nodes from layer $0$ to $i$, and the IPM is used to check the priority of $q_{1:f}$ in turn until it reaches the maximum layer $f$ or quits when the AV's priority is lower than that of other participants (Algo. \ref{alg:ipd_process}, Line 4):
\begin{equation}
    \exists P_i(\Delta t<0 | q^i, V_x) < C_p, \forall V_x \in \mathcal{A}_i,
\end{equation}
\noindent where $\mathcal{A}_i$ is the set of agents who intend to pass the interaction point at layer $i$, $P_i$ denotes the probability in (\ref{eq:ipm_priority}), and $C_p$ is a constant threshold. In the implementation, protection times $\Delta t_p^-$ and $\Delta t_p^+$ are calculated by setting $\Delta v$ to zero since it is hard to estimate $\Delta v$ accurately. Moreover, $\min(t_e^i)$ in $M^-$ uses the $t$ value in $q^i$, and $M^+$ is calculated via the constant acceleration model.
Following the illustration in Fig. \ref{subfig:speed_profile_gen}, once $q_f$ has a lower priority, nodes at previous layers (e.g., $s_{f-1}$) will be checked to see if they meet the following condition
\begin{equation}
\begin{aligned}
    (s, v, t)_{l} &\in S(t_{l}), l < f, 
\end{aligned}
\label{eq:cond_invariable_safe_set}
\end{equation}
\noindent where $S(\cdot)$ is the invariable safe set \cite{pek2020tro_fail_safe} that at least one
action exists for the AV to remain safe for an infinite time horizon. Our work involves two kinds of invariable safe sets. One is to promise the AV can stop before the intersection (Time-to-Brake \cite{hillenbrand2006ttr} greater than 0), and another is to obey the responsibility safety criterion \cite{koopman2019RSS} when following a car. Then, a one-step forward expansion will be conducted at $s_{f-1}$, and the generated child should obey its parent's constraint (\ref{eq:cond_invariable_safe_set}). For example, as the green edges show in Fig. \ref{subfig:speed_profile_gen}, child nodes at the new layer $s_{f^*}$ are sampled if they allow the AV to stop before a safe distance to the intersection.

\begin{figure}
\begin{algorithm}[H]
\caption{Priority Determination}\label{alg:ipd_process}
\begin{algorithmic}[1]
    \State \textbf{Notation}: agent list $\mathcal{A}$

    \State \textbf{Initialize}: $\mathcal{V}, q_{1:f} \gets \textsc{StGraphSearch}()$ 
    \For{i in $\{1,\dots, f\}$}
        \If{$\textsc{PriorityIsLow}(q^i, \mathcal{A}_i)$}
            \State $q_{1:f^*} \gets \textsc{CheckAndExpand}(\mathcal{V}_{0:i-1})$
            \State \textbf{break}
        \EndIf
    \EndFor
    \If{$q_{1:f^*} == \textsc{None}$}
        \State \textbf{return} $\textsc{Smooth}(q_{1:f})$
    \EndIf
    \State \textbf{return} $\textsc{PriorityDetermination}(\mathcal{V}, q_{1:f^*})$
\end{algorithmic}
\end{algorithm}
\vspace{-2em}
\end{figure}


Finally, the best speed profile $q_{1:{f^*}}$ is picked up, and the above operations are executed iteratively (Algo. \ref{alg:ipd_process}, Line 12) until all nodes of the speed profile have a higher interaction priority. Fig. \ref{subfig:speed_profile_gen} shows three speed profiles in orange, purple and grey, respectively. The orange one is the best speed profile derived from the forward simulation while having a low priority at $p_2$. As a result, the speed profile in purple is selected instead. Then, the grey speed profile will be applied if the purple speed profile has a low priority at $p_1$.


\subsection{Speed Profile Smoothing}

The chosen speed profile $q_{1:{f^*}}$ can be smoothed further to meet the kinodynamic feasibility using a quintic piecewise Bézier curve in the spatio-temporal semantic corridor (SSC \cite{ding2019safe}). Based on the $\{s_1, s_2, ..., s_k\}$ and speed limits in the search space, our method use $q_{1:{f^*}}$ as the initial seeds to build SSC, while dynamic obstacles are not mapped to the SSC because the initial seeds have guaranteed the AV's safety. 

\section{Experimental Results}\label{sect:experiments}

We train and test the IPM in the INTERACTION dataset \cite{zhan2019interaction} because it contains motions of various traffic participants in various driving scenarios without traffic lights. The proposed planner is validated in the CARLA \cite{dosovitskiy2017carla}.

\subsection{Implementation Details \label{subsect:experi_imple_details}}

We use Pytorch \cite{paszke2019pytorch} to train the MLP network in IPM, and the whole planning framework is implemented with C++ backend. All experiments are conducted on a computer with an AMD-5600X CPU (3.7 GHz).
From the INTERACTION dataset \cite{zhan2019interaction}, we first extract the data that have coincident trajectories between agents (as plotted in Fig. \ref{fig:illus_interaction_point}) and divide them into different pairs of interactions. Some examples are available on our project website, where agents are simplified as circles. At the same interaction point, agent $m$ overtaking $n$ is identical to the case where agent $n$ gives way to $m$. As these two cases are essentially the same scenario viewed from different perspectives, we only add this type of data to our dataset once to avoid data duplication.

In the simulation, the AV's acceleration limits are $(-3.0, 1.5)$ $\text{m/s}^2$, and its field of view is limited, where other traffic agents will not be observed if their distance to the AV is greater than $120 \cdot \max(1.0 - 2\cdot |\arctan(\Delta y, \Delta x)|/\pi, 0.2)$ $\text{m}$. In the IPM, the MLP network and parameters $\mathbf{C_{(\cdot)}}$ in (\ref{eq:fitting_curves}) are available on our project website. For each intersection with a limited field of view, we apply the heuristic function $\text{Dist}(p_o,p_j) = 0.15 \times L$ to roughly estimate the speed limit in (\ref{eq:speed_cons_obs_point}), where $L$ is the length of the lane in the intersection. During forward simulation, the maximal distance between two sampled path points is $10m$. In Algorithm \ref{alg:s-t_graph_search}, the expansion conditions $t_h=10\text{ s}$, $c_v=0.1$ $\text{m/s}$, grid resolutions $R_v = 0.2$ $\text{m/s}$, and $R_t = 0.2$ $\text{s}$. In the classifier (\ref{eq:ipm_priority}), $C_r^1=0.5$, $C_r^2=1.5$, $C_r^3=5.0$, and $C_p = 0.95$.

\begin{figure}[t]
    \vspace{12pt} 
    \centering
    \subfloat[The gap time $\Delta t$ distribution.]{
        \includegraphics[width=2.8in]{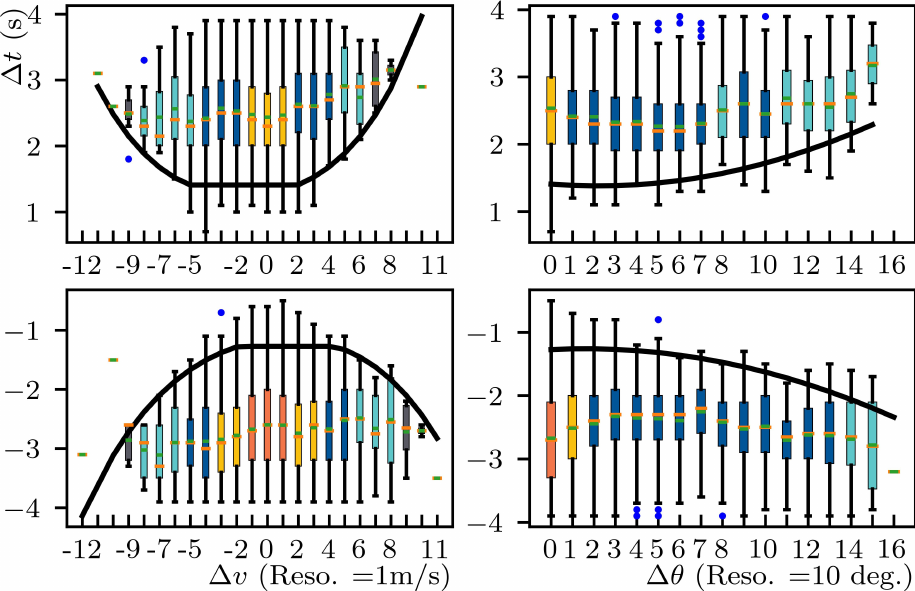}
        \label{fig:interaction_dt_dist}}

    \centering
    \subfloat[MLP training result.]{
        \hspace{-0.2in}
        \includegraphics[width=2.8in]{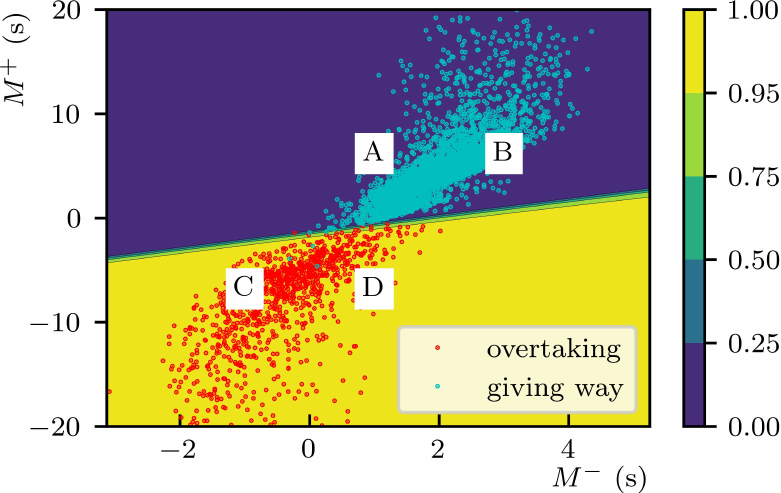}
        \label{fig:priority_net_result}}

    \centering
    \subfloat[Speed boundaries.]{
        \hspace{-0.15in}
        \includegraphics[width=2.8in]{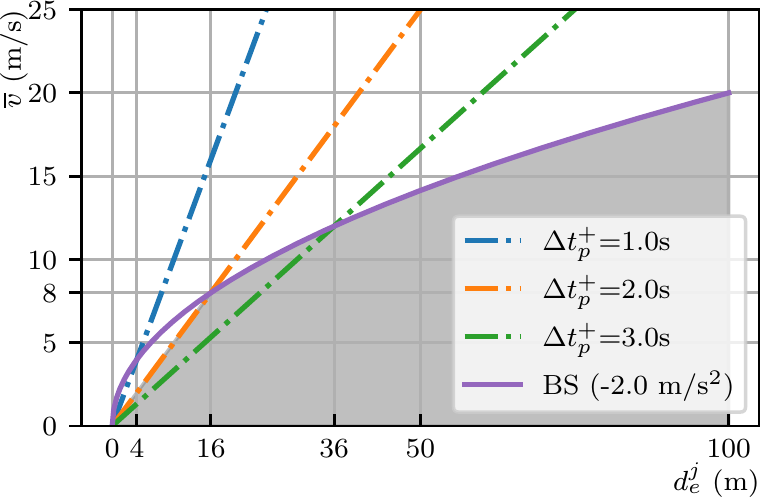}
        \label{fig:speed_cons_compares}}
    \caption{Experiments of the interaction point model. (a) Illustrations of $P(\Delta t | \Delta v$), $P(\Delta t | \Delta \theta$) and fitting curves. (b) Training result of the interaction priority network, where points are labelled samples in the training set, and the background color denotes the probability calculated by \textsc{MLP}. (c) Comparisons of speed limits on different considerations.}
    \label{fig:expri_interaction_dataset}
    \vspace{-1.0em}
\end{figure}

\begin{table}[t!]
    \vspace{2.0em}
    \begin{threeparttable}
        \caption{Time Consumption of Planning Framework}
        \centering
        \label{tab:experi_real_time_per}
        \setlength\tabcolsep{5pt} 
        \begin{tabularx}{\columnwidth}{@{}cccccc@{}}
        \toprule
        Time & $<$1ms & 1-10ms & 10-20ms & 20-30ms & 30-40ms\\
        \midrule
        Percentage & 10.97\% & 42.53\%  & 39.31\% & 7.16\% & $<$0.03\% \\
        \bottomrule
        \end{tabularx}
    \end{threeparttable}
\vspace{-1.5em}
\end{table}

\subsection{Interaction Protection Time \label{subsect:experi_ipm_protect_time}}
\begin{figure*}[tb]
    \vspace{8pt} 
    \centering
    \subfloat[Giving way to other vehicles. (1) Due to the speed limit at $p_o$, the AV slows down from $\text{5.1m/s}$, and it has a lower priority than $\text{V}_{537}$ at both $p_1$ and $p_2$. (2) When $\text{V}_{537}$ turns right, the AV decides to occupy $p_1$. (3) Finally, the AV follows $\text{V}_{537}$ while maintaining a certain speed.]{
        \includegraphics[width=2.2in]{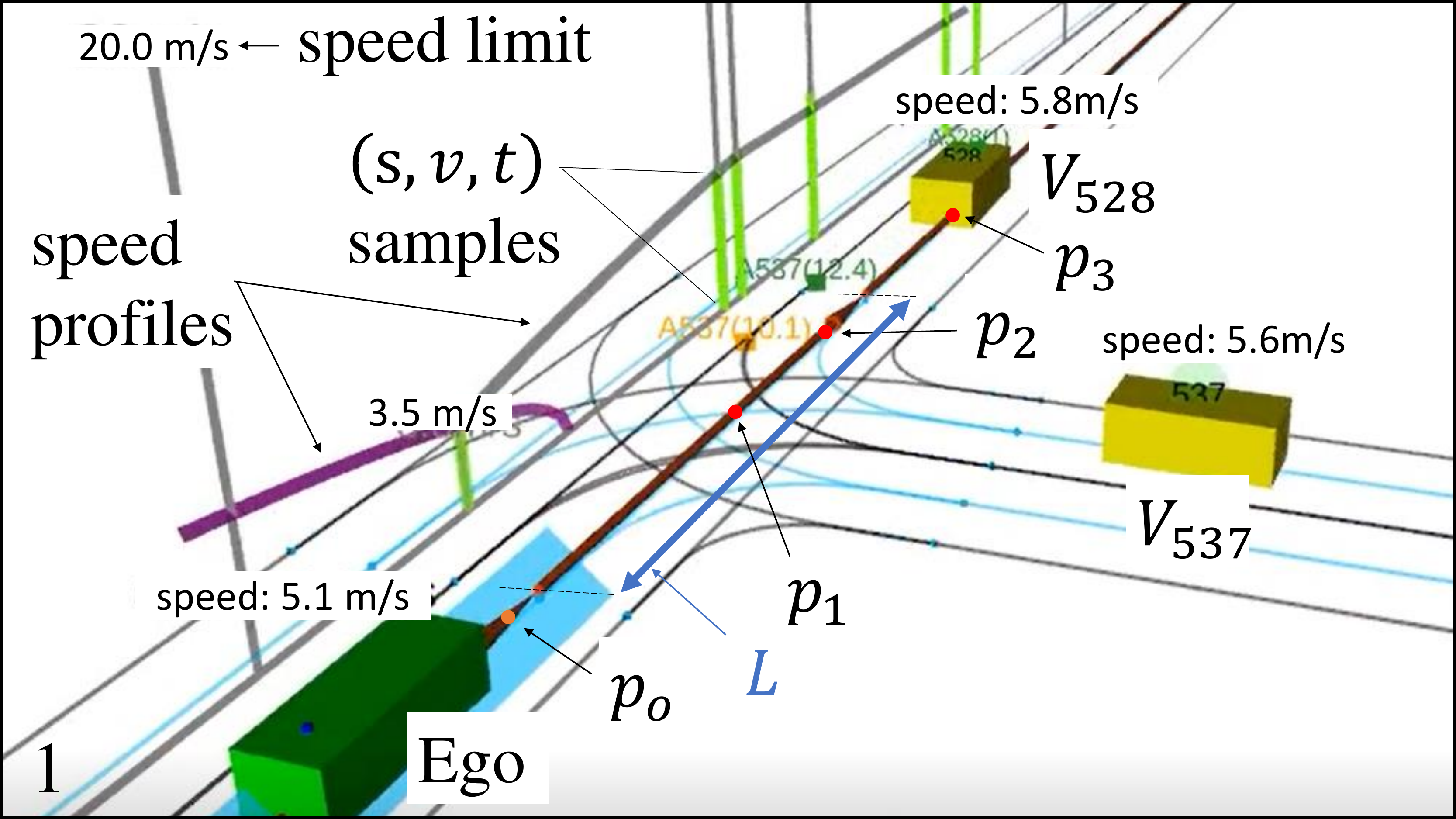}
        \includegraphics[width=2.2in]{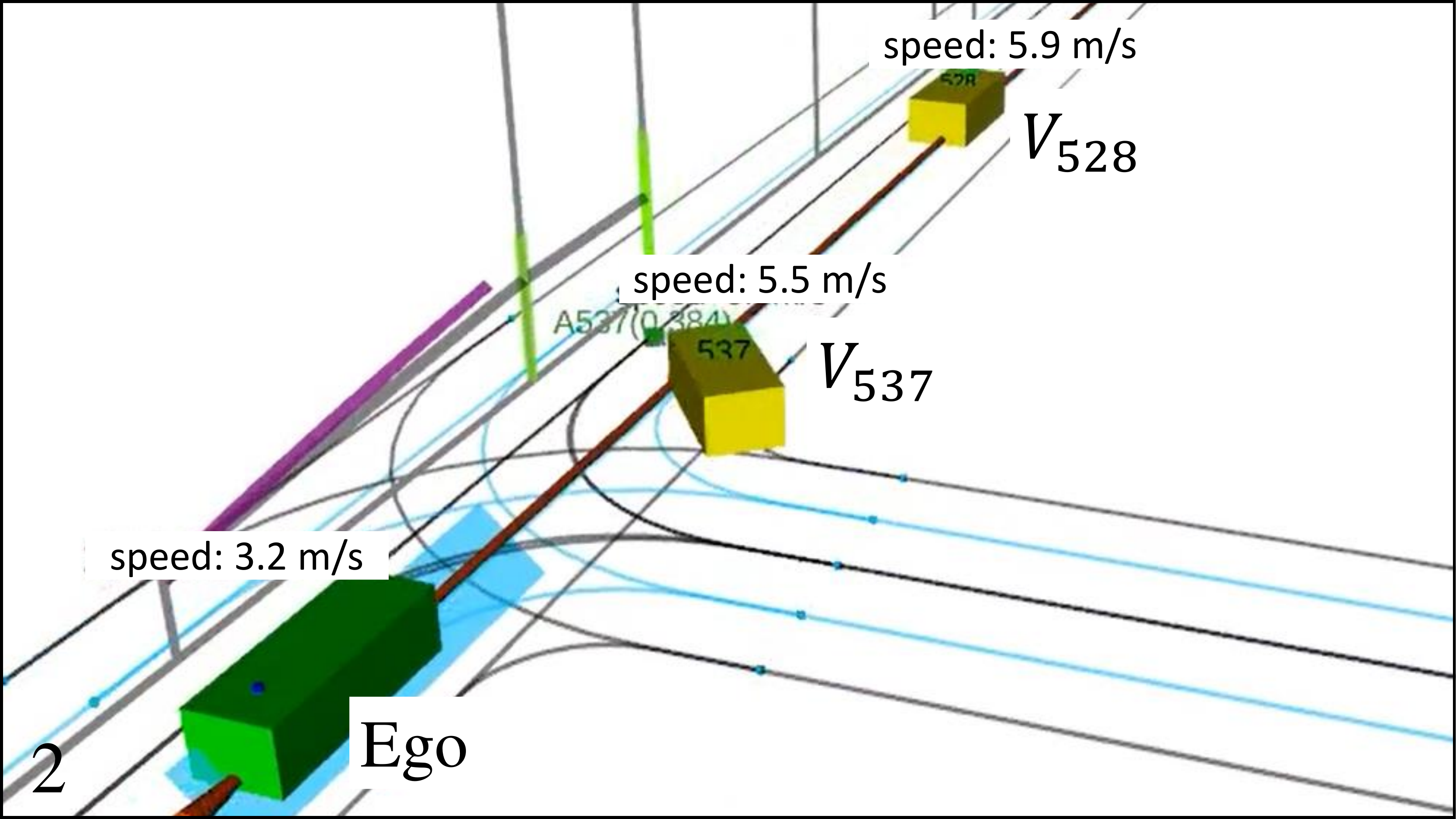}
        \includegraphics[width=2.2in]{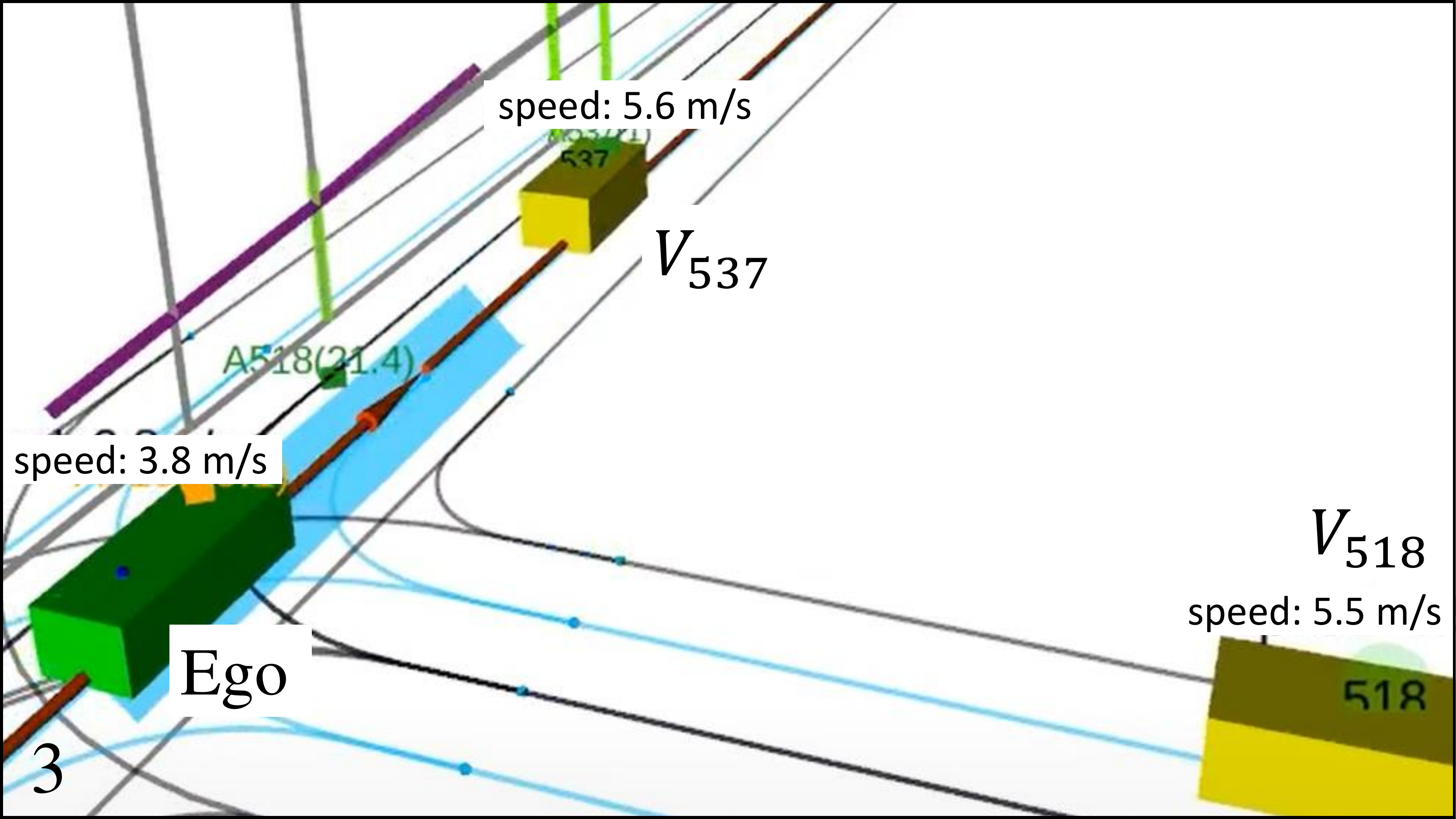}
        \label{subfig:carla_demo2}}

    \subfloat[Overtaking other vehicles. (1) Firstly, the AV decides to stop before $p_1$. (2) Once $\text{V}_{1605}$ turns left, the AV starts to compete with $V_{1069}$. Meanwhile, the AV believes it has a higher priority than $\text{V}_{1609}$ at $p_1$ and $p_2$. (3) The AV successfully overtakes $\text{V}_{1069}$.]{
        \includegraphics[width=2.2in]{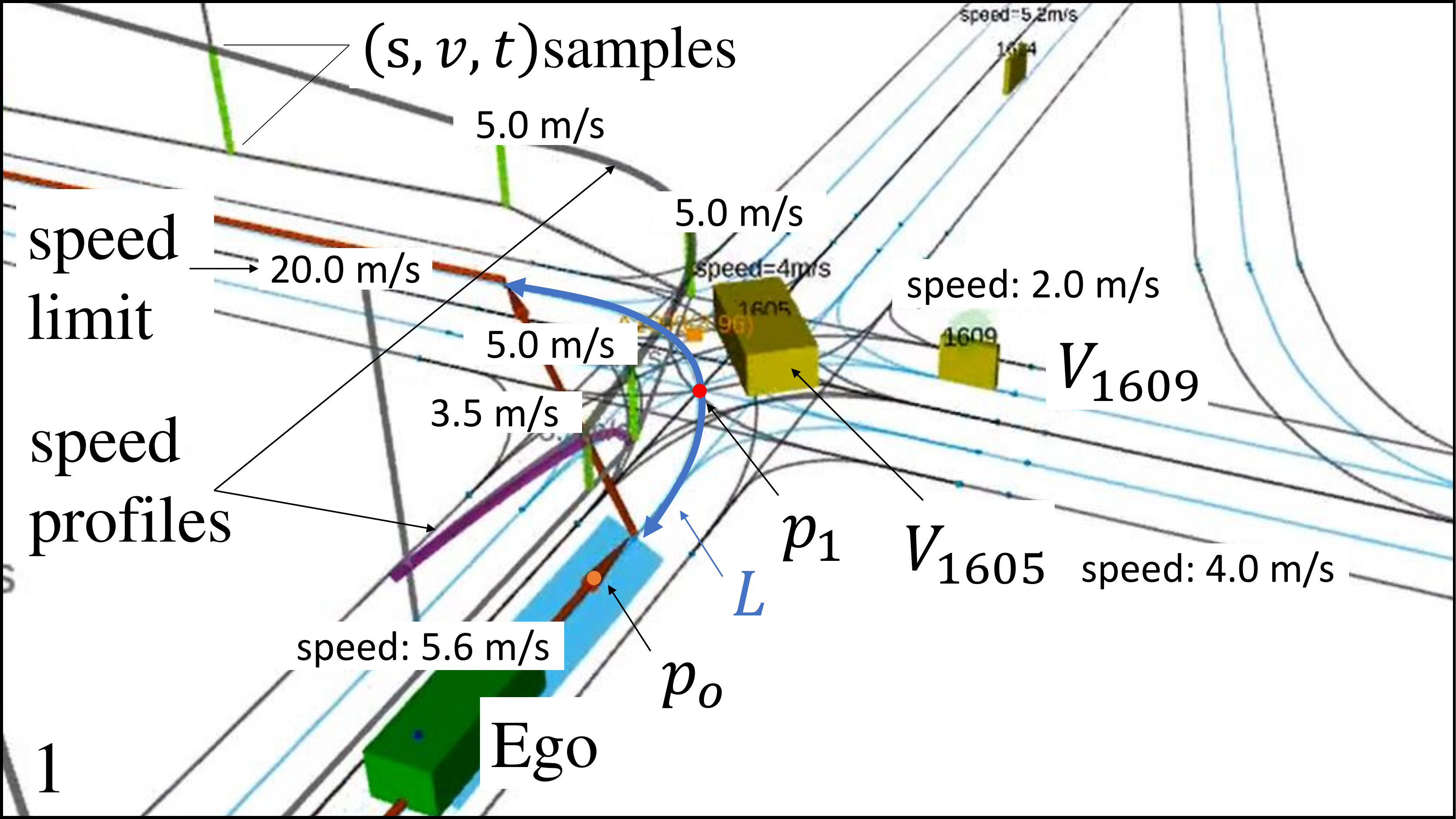}
        \includegraphics[width=2.2in]{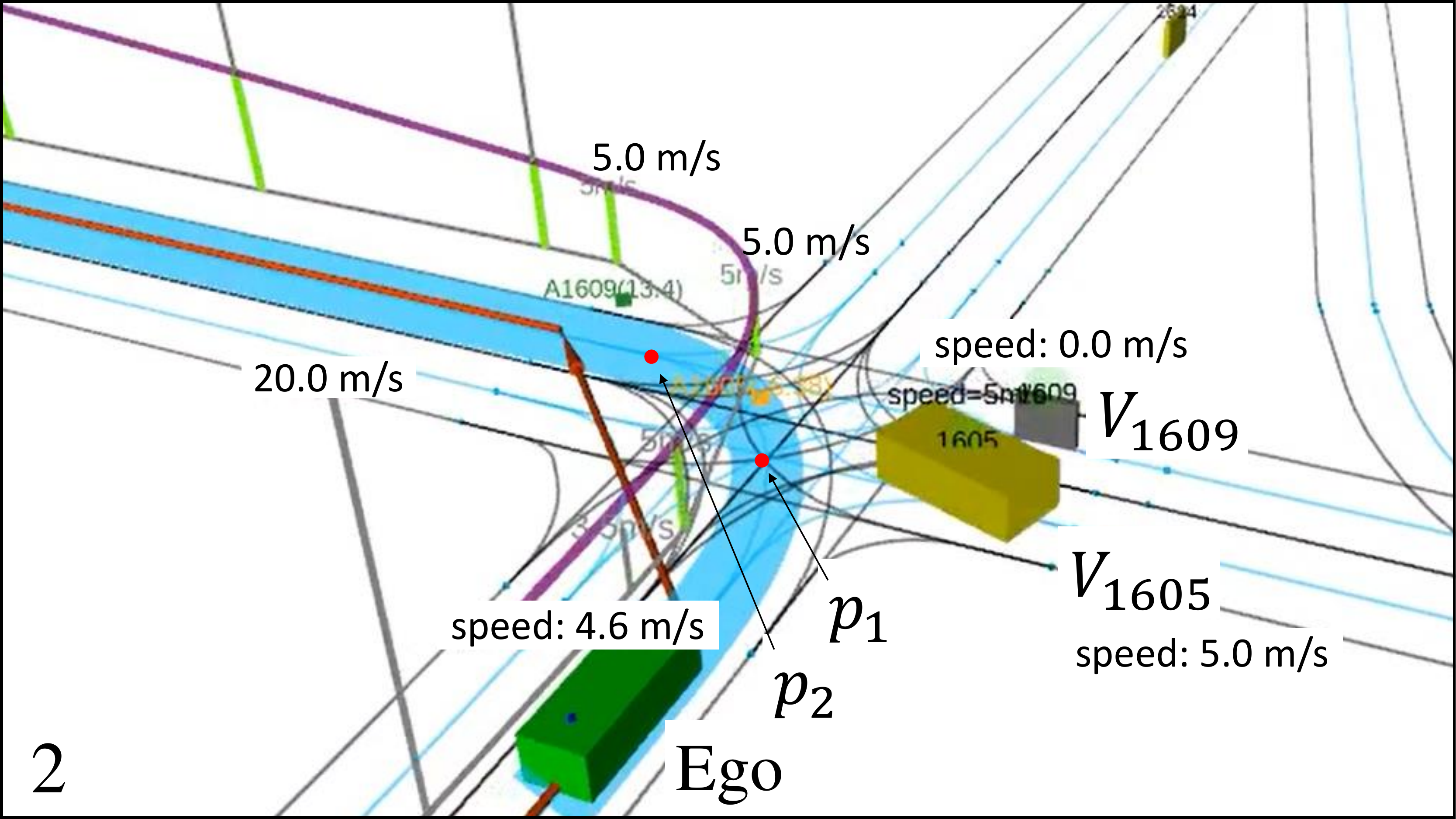}
        \includegraphics[width=2.2in]{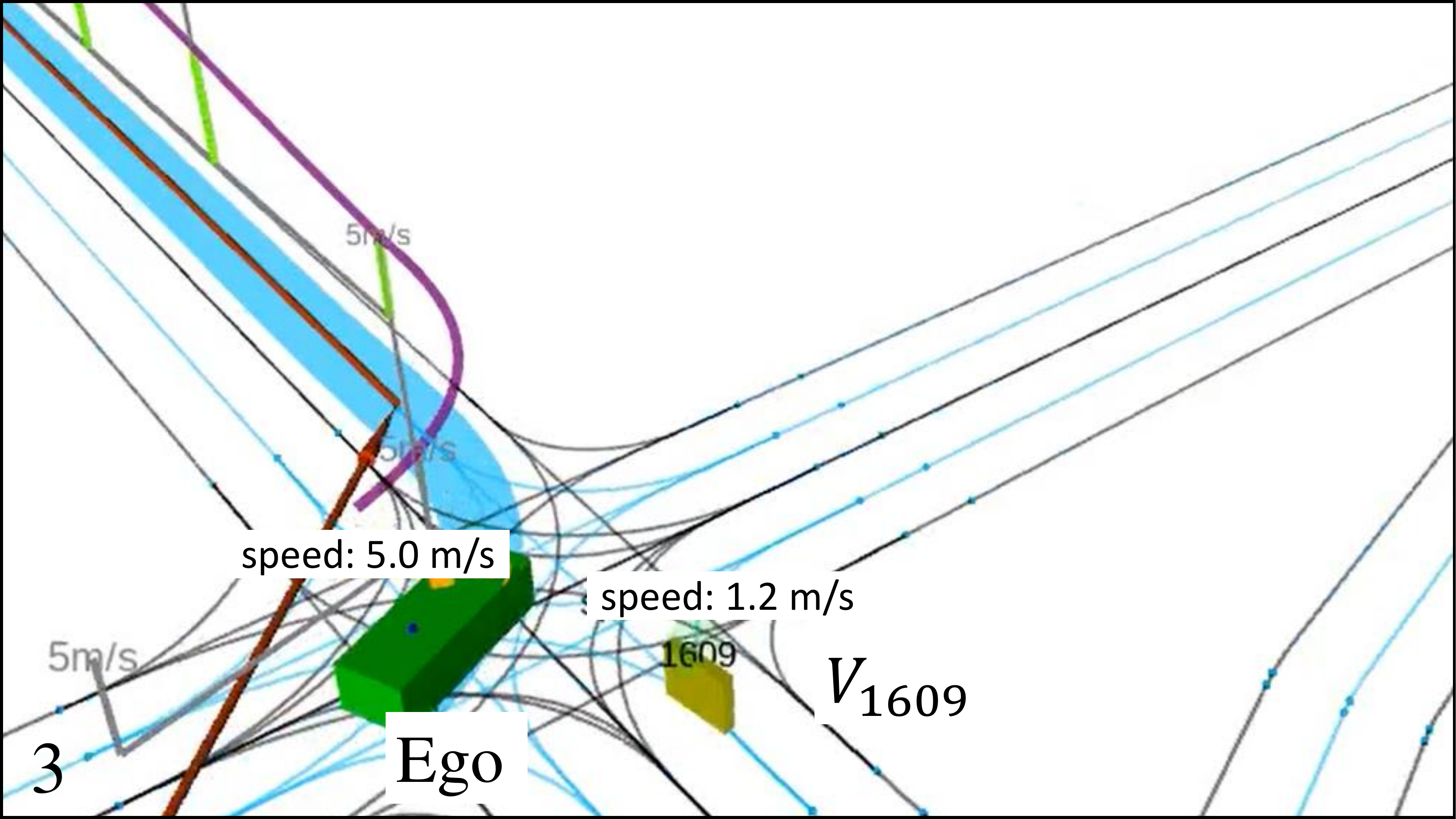}

        \label{subfig:carla_demo1}}

    \caption{Illustration of different interaction results in a conflict zone.}
    \label{fig:carla_demos}
\vspace{-0.5em}
\end{figure*}

\renewcommand{\arraystretch}{1.2} 
\begin{table*}[ht!]
\caption{Quantitative Experimental Results}
\centering
\vspace{-0.75em}
\begin{threeparttable}
\begin{tabular}{lccc|ccc}
\Xhline{2\arrayrulewidth}
& \multicolumn{3}{c|}{Town02 (Seed 1)}                                                                                                                & \multicolumn{3}{c}{Town02 (Seed 2)}                                                \\ \hline
Methods                     & Completion (m) $\uparrow$      & Collision Num.$\downarrow$      & Avg. Speed\tnote{a} (m/s)$\uparrow$       & Completion (m) $\uparrow$     & Collision Num.$\downarrow$   & Avg. Speed (m/s)$\uparrow$     \\ \hline
SSC\cite{ding2019safe}      & 1232.00 $\pm$ 187.71           & \textbf{0.0} $\pm$ 0.00          & 4.10 $\pm$ 0.61           & 1444.94 $\pm$ 081.07          & 0.2 $\pm$ 0.45                & 4.8 $\pm$ 0.25          \\ 
MCTS\cite{hanna2021pomdp}   & 1295.73 $\pm$ 054.82           & \textbf{0.0} $\pm$ 0.00                   & 4.43 $\pm$ 0.31           & 1401.86 $\pm$ 104.27          & \textbf{0.0} $\pm$ 0.00                & 4.7 $\pm$ 0.36          \\ 
MMFN-Expert\cite{mmfnzhang} & 1340.84 $\pm$ 412.39           & \textcolor{red}{1.0} $\pm$ 0.71          & \textbf{4.86} $\pm$ 0.53           & \textbf{1620.00} $\pm$ 059.71 & \textcolor{red}{0.8} $\pm$ 0.84                & \textbf{5.4} $\pm$ 0.18 \\ 
PD-$M^-$                    & 1193.86 $\pm$ 107.04           & \textbf{0.0} $\pm$ 0.00          & 3.98 $\pm$ 0.38           & 1303.22 $\pm$ 035.57          & \textbf{0.0} $\pm$ 0.00       & 4.3 $\pm$ 0.11          \\ 
PD-CVel                     & 1303.08 $\pm$ 161.48           & 0.3 $\pm$ 0.48                   & 4.41 $\pm$ 0.52           & 1398.94 $\pm$ 081.53          & 0.2 $\pm$ 0.45                & 4.7 $\pm$ 0.30          \\ 
\rowcolor[HTML]{E7E6E6}
PD-IPM                      & \textbf{1358.11} $\pm$ 060.73  & \textbf{0.0} $\pm$ 0.00          & 4.71 $\pm$ 0.48  & 1457.96 $\pm$ 042.54          & \textbf{0.0} $\pm$ 0.00       & 4.9 $\pm$ 0.13\\
\Xhline{2\arrayrulewidth}
\end{tabular}

\smallskip
\scriptsize
\begin{tablenotes}
\RaggedRight
\item[a] is the average velocity before the AV collides with other vehicles and cannot continue with the navigation task.\\
\end{tablenotes}
\end{threeparttable}
\label{tab:experi_nav_tasks}
\vspace{-1.5em}
\end{table*}
\renewcommand{\arraystretch}{1.0}

We analyze the extracted pairs of interactions (mentioned in Sect. \ref{subsect:experi_imple_details}) and plot $P(\Delta t | \Delta \theta)$ and $P(\Delta t | \Delta v)$ in Fig. \ref{fig:interaction_dt_dist}. Inside a box, the green line represents the mean value, and the orange line denotes the median. The color of the box represents the sample size, where orange means the size is greater than 10000, yellow (than 1000), blue (than 100), light blue (than 10), and black is [1,10). In addition, the solid lines are fitting curves of $\Delta t_p^-(\cdot)$ and $\Delta t_p^+(\cdot)$ in (\ref{eq:fitting_curves}), which are based on the Gaussian $3\sigma$ intervals.

\subsection{Interaction Priority \label{subsect:experi_ipm_priority}}

As shown in Fig. \ref{fig:priority_net_result}, we label samples with negative $\Delta t$ as overtaking (red points) and positive as giving way (cyan points) in the dataset. 
The background color denotes $\textsc{MLP}(\Delta t < 0 | M^-, M^+)$ in (\ref{eq:ipm_priority}),
and there are four areas.
For B and C, the priority is clear that can be quickly classified. In A and D, $M^+$ plays a key role. Most agents in the dataset can give way to their competitors when the giving way ability is higher.
We further classify $\textsc{MLP}(\Delta t < 0 | M^-, M^+) \geq 0.5$ as overtaking to evaluate the prediction accuracy. The trained $\textsc{MLP}$ network achieves 99.79\% accuracy in the training set with a data size of 5283 and 99.81\% in the validation set (10071 data size). In the validation experiment, we test all time slots before reaching the interaction point, which makes the validation set's size larger than the training set's.

\subsection{Advance Deceleration for Emergency \label{subsect:experi_ipm_speed_cons}}

Besides the response time factor, the IPM explains why an advance deceleration is necessary for an emergency from another perspective. As shown in Fig. \ref{fig:speed_cons_compares}, the purple line represents the braking speed limit, which is the maximal speed when the braking distance is less than $d_e^j$, and the dashed lines are the speed limits defined by (\ref{eq:speed_cons_obs_point}). This figure suggests that when $\Delta t_p^+$ is $2.0$ s and the AV runs at $8.0$ m/s, the braking is hard to prevent a collision if a vehicle suddenly appears within $16$ m. That is, it reflects the speed limits of the grey area.

\subsection{Qualitative Results \label{subsect:experi_qualti_result}}


To verify that the proposed planning framework is competent in highly interactive scenarios, we test our method in Town02 and Town04 of CARLA. In the experiments, traffic agents in the simulation are set to ignore the traffic signals and signs, and two simulation results are discussed in Fig. \ref{fig:carla_demos}. In addition, more qualitative results are available in the supplementary video on the project website.

\subsection{Quantitative Results \label{subsect:experi_quanti_result}}

1) \textit{Real-time Performance}: A 30-minute navigation test with dynamical agents is conducted in CARLA to verify the computational efficiency of our proposed planning framework. The test results are shown in Table \ref{tab:experi_real_time_per}, where the computation time includes all planning modules in Fig. \ref{fig:overview-framework}, and the average calculation time is $10.24$ ms. We observe that in 92.8\% of cases, the runtime of our proposed method is less than $20$ ms, which shows the real-time performance.

2) \textit{Performance Comparisons}: We select Town02 as the test scenario because the map is relatively small and more suitable for comparing the interaction performance. In addition, the position of interaction points in Town02 is fixed so that we do not need to predict other agents' paths in advance. Like previous experiments, all junctions and intersections on the map are unprotected by the traffic rules. The following algorithms are compared with the same initial location and different random seeds.
\textbf{SSC} \cite{ding2019safe}: constant velocity (CVel) model is applied to generate the SSC. \textbf{MCTS} \cite{hanna2021pomdp}: Monte Carlo tree search algorithm, where other agents are predicted to slow down to stop or speed up to a certain speed when tracking the path, and their goal probabilities are uniform. \textbf{MMFN-Expert} \cite{mmfnzhang}: an elaborated expert agent. \textbf{PD-M}$^{-}$: this method is based on the proposed planning framework, but the priority determination module decides to overtake when $M^-$ is less than zero. \textbf{PD-CVel}: like PD-$M^{-}$, it overtakes when the $\Delta t$ estimated by CVel is less than zero. \textbf{PD-IPM}: the proposed planning framework using (\ref{eq:ipm_priority}) to determine the priority. The speed constraint (\ref{eq:speed_cons_obs_point}) is enforced in all methods since most benchmarks \cite{ding2019safe, hanna2021pomdp} did not model the occluded factors.

Each algorithm is tested with a series of goals in five minutes. The tests are repeated five times, and the performances are recorded in Table \ref{tab:experi_nav_tasks}. In the table, the numbers are the mean values and standard deviations for three metrics: completion distance, number of collisions, and average speed.
The experimental results demonstrate that our method is safe and efficient. In the seed 1 setting, PD-IPM achieves no collision situations with the longest completion length. In the seed 2 setting, although the MMFN-Expert has the longest length and highest average speed, it is unsafe and has a high collision number in red. By contrast, our PD-IPM gets a zero-collision rate and outperforms other approaches \cite{ding2019safe, hanna2021pomdp}. Besides this, the performance of PD-$M^-$ is conservative but safe, which shows our planning framework's reliability. The comparison between PD-CVel and PD-IPM reflects that the IPM can make the planning system more effective.


\section{CONCLUSIONS \label{sect:conclusion}}

This paper presented a novel unified planning framework based on the proposed interaction point model (IPM), which enabled a uniform description of the interaction under various driving scenarios. First, the IPM was trained and tested in the real traffic data, showing the general interaction rule in different scenarios. Next, the planning framework was introduced, and the efficient graph search and interaction priority determination were elaborated. Experiments showed that our framework enabled the system to interact safely with other agents, and the IPM made it more efficient.
Future work will include introducing more complicated scenarios into the planning framework, with interactions in open space, with pedestrians, or between other agents under more traffic regulations modeled.






\bibliographystyle{IEEEtran}
\bibliography{IEEEabrv,ref}

\end{document}